\documentclass[letterpaper, 10 pt, conference]{ieeeconf}  

\IEEEoverridecommandlockouts                              

\usepackage{graphics} 
\usepackage{graphicx}
\usepackage{epsfig} 
\usepackage{amsmath} 
\usepackage{amssymb}  
\usepackage[english]{babel}
\title{\LARGE \bf
Learning Bayes Filter Models for Tactile Localization
}

\author{Tar{\i}k Kele\c{s}temur$^{1}$, Colin Keil$^{2}$, John P. Whitney$^{1}$, Robert Platt$^{2}$ and Ta\c{s}k{\i}n Pad{\i}r$^{1}$
\thanks{*This material is based upon work supported by the National Science Foundation under Award No. 1928654, 1724257, 1724191, 1750649, 1763878, 1830425, the Office of the Secretary of Defense under Agreement Number W911NF-17-3-0004, and the U.S. Office of Naval Research under award number N00014-19-1-2131.
}
\thanks{$^{1}$Tar{\i}k Kele\c{s}temur, $^{1}$John P. Whitney, $^{1}$and Ta\c{s}k{\i}n Pad{\i}r are with the College of Engineering, and $^{2}$Colin Keil and $^{2}$Robert Platt are with the Khoury College of Computer Sciences, Northeastern University, Boston, Massauchusetts 02115, USA. 
{\tt\small{kelestemur.t@northeastern.edu}}
}%
}

\begin{document}

\maketitle
\thispagestyle{empty}
\pagestyle{empty}

\begin{abstract}
Localizing and tracking the pose of robotic grippers are necessary skills for manipulation tasks. However, the manipulators with imprecise kinematic models (e.g. low-cost arms) or manipulators with unknown world coordinates (e.g. poor camera-arm calibration) cannot locate the gripper with respect to the world. In these circumstances, we can leverage tactile feedback between the gripper and the environment. In this paper, we present learnable Bayes filter models that can localize robotic grippers using tactile feedback. We propose a novel observation model that conditions the tactile feedback on visual maps of the environment along with a motion model to recursively estimate the gripper's location. Our models are trained in simulation with self-supervision and transferred to the real world. Our method is evaluated on a tabletop localization task in which the gripper interacts with objects. We report results in simulation and on a real robot, generalizing over different sizes, shapes, and configurations of the objects.
\end{abstract}
\section{INTRODUCTION}
Humans take advantage of tactile perception in many manipulation tasks. We can locate, recognize, grasp, and manipulate objects without looking at them. Tactile perception is also very useful in situations where visual feedback is not available. For example, we can easily find our wallet on a table in a dark room. The ability to make sense of tactile input has influenced neuroscience researchers. In the early work~\cite{bach1969vision}, the researchers coined the term \textit{sensory substitution} in which humans are able to use one sensory modality, e.g., tactile, to stimulate another modality, e.g., vision. This and following work has led to the invention of devices that lets visually-impaired people to `perceive' the world with touch. Moreover,~\cite{takahashi2009integration} showed that humans integrate vision and touch when they are using tools in a near-optimal fashion.

Tactile perception is a desired skill for several robotic applications. Similar to humans, the studies have shown that the robots that can leverage touch feedback are more effective than those that only rely on visual or proprioceptive sensors. Some of the examples where tactile perception is shown to be essential are grasping~\cite{merzic2019leveraging}, shape reconstruction~\cite{smith20203d}, contact-rich tasks~\cite{lee2019making}, and object recognition~\cite{lin2019learning}. One other manipulation skill that benefits from the tactile sensing is tactile localization where the aim is to estimate the pose of an object or an end-effector using tactile feedback received from the environment. While the majority of the earlier work aims to localize the objects (as discussed in Section~\ref{sec:related}), we focus on the problem of localizing the gripper with respect to a global map of the environment. Compared to the localization with visual feedback~\cite{cifuentes2016probabilistic}, tactile localization is not affected by the occlusions between the camera and the end-effector. Tactile localization can be useful in situations where there is no precise forward kinematic model of a robotic arm (e.g. low-cost manipulators) or the pose of the arm cannot be defined in with respect to the world coordinates.
\begin{figure}[!t]
    \centering
    \includegraphics[trim={1cm 0cm 1cm 0cm}, clip, width=0.98\columnwidth]{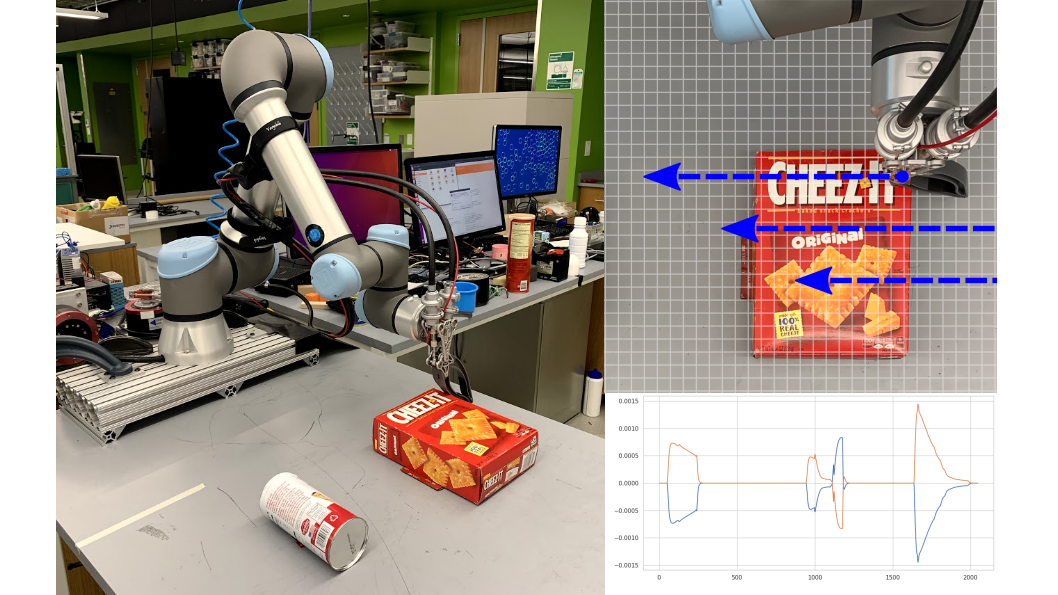}
    \caption{\textbf{Tactile Localization Task:} The gripper traverse over the table and objects (top-right) while collecting tactile observations (bottom-right) to localize itself.} \label{fig:real}
\end{figure} 

Localization is a fundamental problem in robotics. The majority of the work focuses on this problem in the context of mobile robot navigation where the goal is to find and track the location of a mobile robot on a global map. Tactile localization is similar to the mobile robot localization in terms of the problem formulation but differs in the sensing modalities. In tactile localization, the robot receives a ``sense of touch'' from the environment in the form of proprioceptive measurements (joint angles and velocities) or external sensor readings (force or torque). These tactile measurements can be used to find which state the robot is in. 

Tactile feedback has a limited use in state estimation due to its noisy and high-frequency nature, but it can provide rich information about the contact interactions, which can help determine object properties and locations. Bayes filters~\cite{thrun2002probabilistic} are well suited for state estimation problems with uncertain observations and transition functions. The Bayes filters work by maintaining a \textit{belief} of the state and use an observation and a motion model. The observation and motion models update the belief using the measurement received from the environment and the actions taken by the robot.  In our case, these models are not known so we propose to learn them from data. We design observation and motion models using neural network layers and implement Bayes filtering as tensor operations. Our observation model is based on the U-net architecture~\cite{ronneberger2015u} and facilitates the multi-modal sensor inputs. It uses the images of the environment as visual maps and tactile feedback as the observation to generate the likelihood probabilities, and the motion model transitions the belief to the next timestep. We evaluate our method on a tabletop localization task in which the robotic gripper traverses over the table and interacts with objects. Our approach is compared against two baselines: a uniform observation model and a naive version of our observation model which does not use the visual maps. The contributions of this paper are to (1) formulate tactile localization as a Bayes filtering problem with learned models, (2) introduce a self-supervised data collection procedure to gather a large amount of contact interaction data to train these models, and (3) show that these models can generalize over novel objects and can be directly transferred into real-world. Our method is able to achieve 93\% success rate in the simulation and 90\% in the robot experiments.

\section{RELATED WORK} \label{sec:related}
\subsection{Learning Bayes Filters}
Model-free learning has shown great success in many robotics tasks in the last few years. Also, the progress in the sim-to-real research made it possible to train large neural networks in simulation and transfer these models to the real world. However, choosing the right class of models is still an open problem for many robotic tasks. One approach that addresses this issue is embedding robotic algorithms as priors into model-free learning.~\cite{Jonschkowski-16-NIPS-WS} proposed a Histogram filter that can be trained end-to-end and showed that it outperforms LSTM networks for state estimation problems.~\cite{karkus2017qmdp} combined the learnable Histogram filters with QMDP method to solve planning under partial observability. The authors concurrently extended their methods to continuous state space by proposing differentiable particle filters~\cite{Jonschkowski-RSS-18, karkus2018particle}.~\cite{singh2018active} and~\cite{gottipati2019deep} combined learnable Bayes filter with deep reinforcement learning to achieve active localization. All the methods mentioned above have conducted their experiments in simulation and focused on the problem of mobile robot localization with visual or lidar sensors except~\cite{karkus2017qmdp} in which the authors also show results on a 2D grasping experiment. In contrast, we focus on the problem of tactile localization where the objective is to estimate the position of a gripper with respect to the global map of the environment.

\subsection{Tactile Localization}
The work of~\cite{gadeyne2001markov, corcoran2010measurement, petrovskaya2006bayesian, saund2017touch} are early examples of using sequential Bayes filters to estimate the object pose using fingertip contact sensing. The focus is on estimating the full pose of a simple object. The work above does not take into account any prior knowledge of the motion dynamics of the object. However, if the robot is manipulating the object, we can potentially leverage the process dynamics to improve object tracking. This idea is explored in~\cite{liang2020hand}. The work in~\cite{pfanne2018fusing} fuses observations of visual features with the touch information. The idea is to use visual information in combination with the touch data to better localize the object. However, the object must be visible to the camera during manipulation – something that can be difficult. In the recent years, high-resolution tactile sensors such as Gelsight have gained growing attention. Early work ~\cite{li2014localization} localized the pose of a USB stick by matching the tactile imprint against a model of a USB symbol using this sensor. Furthermore,~\cite{li2014localization} fused the Gelsight data with the point cloud using the Iterative Closest Point (ICP) algorithm.~\cite{bauza2019tactile} also uses ICP for localization with pre-computed tactile shape. Recently,~\cite{Wirnshofer-RSS-20} used Particle filters for localization of objects using the proprioceptive sensors of a robotic arm. The belief of the objects is used as the input to the reinforcement learning to guide the actions. This work focus on the local localization problem where the robot has a non-uniform prior as the initial belief. Another limitation of this work is that the pose of the arm is assumed to be known with respect to the world coordinates. The objective of all the work described above is to localize the pose of an object with respect to the gripper. The dual of this problem is to localize the gripper with respect to the environment. For example, the work in~\cite{platt2011using} localizes the pose of a robotic hand with respect to a flexible piece of plastic textured to the environment. Similarly,~\cite{luo2015localizing} propose to use a visual-tactile sensor to match the features of the sensor reading with the pre-generated features of a fixed environment image to localize the gripper. The feture matching is done by scale-invariant feature transform (SIFT) method. In both of these work, the environment is fixed and known beforehand. Our method is able to generalize over different environment configurations and does not use any feature pre-computation.

\section{PROBLEM STATEMENT} \label{sec:prob}
Bayes filters are a family of algorithms used for state estimation problems using the observations received from the environment and actions taken by the agent. A Bayes filter estimates the state of a dynamical system by maintaining a posterior probability over states that conditions on observations and actions. Let $s_t$ be the state, $o_t$ be the observation, and $a_t$ be the action of the system at time \textit{t}. The posterior probability (also called \textit{belief}) is then written as : $bel(s_t) = p(s_t| a_{1:t-1}, o_{1:t})$. The Bayes filters updates the belief by taking the \textit{prediction} step and \textit{observation update} step:

\begin{equation*}
    bel(s_t) = \underbrace{\vphantom{\sum_{s_{t-1} \in  S}p(s_t | s_{t-1}, a_{t-1})bel(s_{t-1})}\eta p(o_t | s_t)}_{\textit{Observation Update}} \underbrace{\sum_{s_{t-1} \in  S}p(s_t | s_{t-1}, a_{t-1})bel(s_{t-1})}_{\textit{Prediction Update}}
\end{equation*}
where $p(o_t | s_t)$ is the observation model, $p(s_t | s_{t-1}, a_{t-1})$ is the motion model and $\eta$ is the normalization factor. If there is no prior information of system's state, the belief at t=0, $bel(s_0)$ can be initialized uniformly. Note that Bayes filters assume to have the Markov property of the states in which the current belief $bel(s_t)$ contains full information of the past observations and actions.
\begin{figure*}[t]
    \centering
    \includegraphics[trim={1cm 6cm 1cm 0.7cm}, clip, width=0.9\linewidth]{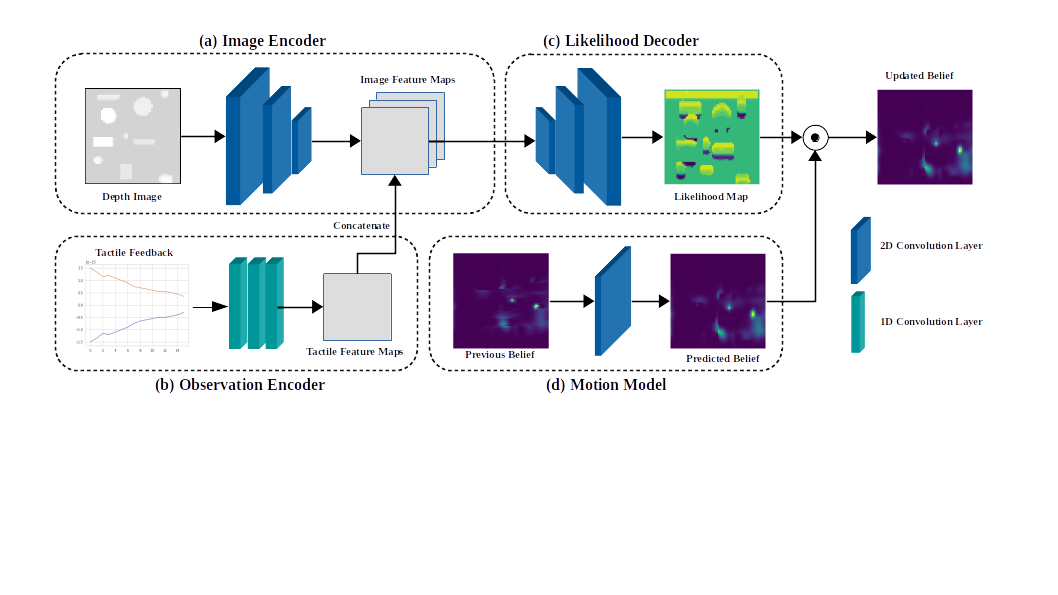}
    \caption{\textbf{Bayes Filter Network Diagram:} The depth image and tactile observation are fed into their encoders (a, b) and the output feature maps are concatenated. The (c) likelihood decoder takes these feature maps and generates the likelihood map. The motion model (d) predicts the belief at next timestep which is then multiplied with the likelihood map and normalized to produce the next belief.}
    \label{fig:network}
\end{figure*}

Our overarching goal is to localize a robotic gripper with respect to a variety of scenes. One possible approach to find out the gripper's location is to visually track it with an external camera, however, this is not feasible due to occlusions by the arm. Furthermore, this method requires a calibration that needs to be reperformed every time the camera or the robot moves.  We, therefore, must rely primarily on touch-based localization during contact interactions. The robot has the opportunity to take a depth image of the environment prior to the interaction but after that, it only has access to the tactile feedback. The tactile observations during the interaction is conditioned on this environment image. The core of our touch localization framework is a discrete Bayes filter (Histogram filter) with learned models. To this end, we have two goals: (1) Learn an observation model that is conditioned on an environment image. This observation model would describe the expected tactile feedback as a function of the gripper position for a new scene, as conveyed by the image of the scene. (2) Learn a motion model that will transition the belief given the action and old state. 

We propose to represent the observation and motion models of the Bayes filter as neural networks and learn them using gradient descent. Let $f_O(\cdot)$ be a neural network that takes the environment image, tactile observation and the action as input and generates the likelihood probabilities of the current observation, i.e. $f_O(o_t, a_t, I) = p(o_t|s_t, a_t, I)$. Let $f_M(\cdot)$ be a neural network that takes the previous belief and the action as input and predicts the belief at the next timestep, i.e. $f_M(bel(s_t-1), a_t) = \overline{bel}(s_t)$. The state space is defined as the projected pixel coordinates of the gripper in the environment image: $s_t = (p_x, p_y) \in \mathcal{Z}^{H\times W}$ where $H$ is the height and $W$ is the width of the image, $I$. The belief is encoded as a $H\times W$ matrix and computed as:
\begin{equation*}
    bel(s_t) = \eta f_O(o_t, a_t, I) \odot f_M(bel(s_t-1), a_t)
\end{equation*}
where $\odot$ is element-wise multiplication. 
\section{LEARNING BAYES FILTER MODELS}
\subsection{Observation Model}
Our observation model is based on the U-net autoencoder architecture introduced in~\cite{ronneberger2015u}. It consist of 3 modules: (a) \textit{Image Encoder}, (b) \textit{Observation Encoder} and (c) \textit{Likelihood Decoder}. The image encoder takes the depth image and passes it through 6 2-D convolution layers. Each layer is followed by batch normalization and ReLu activation. We apply max-pooling after the 4th and the 6th layers.  Similarly, the observation encoder feeds the tactile observations into 3 1-D convolution layers where the first two layers are followed by ReLu activation. The output of the observation encoder is reshaped into a 2D matrix with the same dimension of the image feature maps. The output of these encoders is then concatenated and fed into the likelihood decoder. We use two transposed convolution layers to upsample the feature maps. After each transposed convolution, there are two 2-D convolution layers which are followed by batch normalization and ReLu activation. The first one has 32 and the second one has 16 filters. Finally, we apply a single 2D convolution layer with N filters where N is the number of discrete probability values. Each pixel gives the independent probability of receiving the current observation. We apply a channel-wise \textit{softmax} normalization and train with categorical cross-entropy loss. 

\subsection{Motion Model}
We define the motion model as a 2D convolution operation where the kernel weights $\omega_M \in R^{3\times 3}$ represent the transition probability function. Since the transition function is probability mass function, the output of the convolution operation needs to be positive and sum to 1. This is enforced by a \textit{softmax} normalization after multiplying the kernel. We use a kernel size of 3 because the gripper can only traverse 1 pixel at each time step. The motion model is trained with the binary cross-entropy loss. 

\begin{figure}[t]
    \centering
    \includegraphics[trim={1cm 0cm 1cm 0cm}, clip, width=\columnwidth]{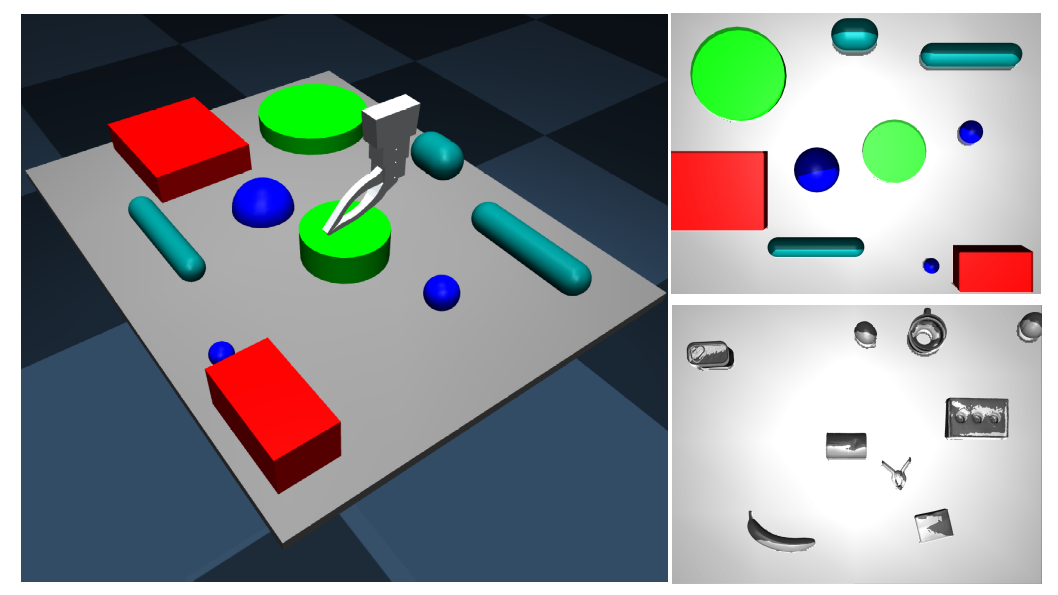}
    \caption{\textbf{Left:} Simulation Environment \textbf{Top-Right:} Training Objects \textbf{Bottom-Right:} Test Objects}
    \label{fig:sim}
\end{figure}
\subsection{Data Collection}
Training neural networks requires a substantial amount of training data. Recent work \cite{lee2019making, ebert2017self} showed that simulations can be used to generate large amount of labeled data by self-supervision. In this setting, the agent explores the environment by following a random policy and collects data from the simulation environments. We follow a similar data collection procedure to train our observation and motion models. A simulation environment is developed using the MuJoCo~\cite{todorov2012mujoco} physics engine to collect tactile observations and their corresponding states. The environment consists of a free-floating gripper, a table and objects on the table (see Figure ~\ref{fig:sim}). A top-down facing depth camera is positioned over the table. The gripper moves from one edge of the table with a linear motion up to the opposing edge of the table. The state space is comprised of the coordinates of the gripper in the camera frame. In order to find the pixel coordinates of the gripper, we first transform the pose of the gripper base to the camera frame and then project it into pixel coordinates: $p = M_{int}M_{ext} P_w$ where $M_{int}$ and $M_{ext}$ are intrinsic and extrinsic camera matrices, respectively, $p=(p_x, p_y)$ is the pixel coordinates of the gripper, and $P_w$ is 3D position of the gripper's base in the environment. 

The joint angles of the gripper are used as the tactile observations. The gripper used in this work has hydro-static linear actuators~\cite{schwarm2019floating} which allows us to set the finger joint stiffness to a low value. This way, the gripper can interact with objects without moving them. To generate the dataset, we first sample objects from our object set. The sizes and the positions of the objects are sampled from a uniform distribution. Then the gripper traverse over every pixel row and column while interacting with the table and objects. The dataset is populated with the depth images, observations, likelihood maps and states: $\mathcal{D} = \{(o_N, s_N, L_N, I)^i\}$ where $i$ is the index of the environment configuration, N is the number of pixels in the image, and $L$ ground-truth likelihood map. Note that we take images of the environment before that episode starts. To train the observation model with full supervision, we need to generate the ground truth likelihood maps for each observation. This likelihood map would describe the probability of receiving current observation given any state. In a single configuration of objects, the gripper collects observations from each state. To generate the likelihood maps, we look at the distance between the current observation and all the other observations. These distances are then normalized between 0 and 1 where the closest distance is 1 and farthest distance is 0 to represent the probability values. Finally, the probability values are discretized into 16 values.
\section{EXPERIMENTS}
\subsection{Simulation Experiments}
\begin{figure*}[t]
    \centering
    \includegraphics[width=0.98\linewidth]{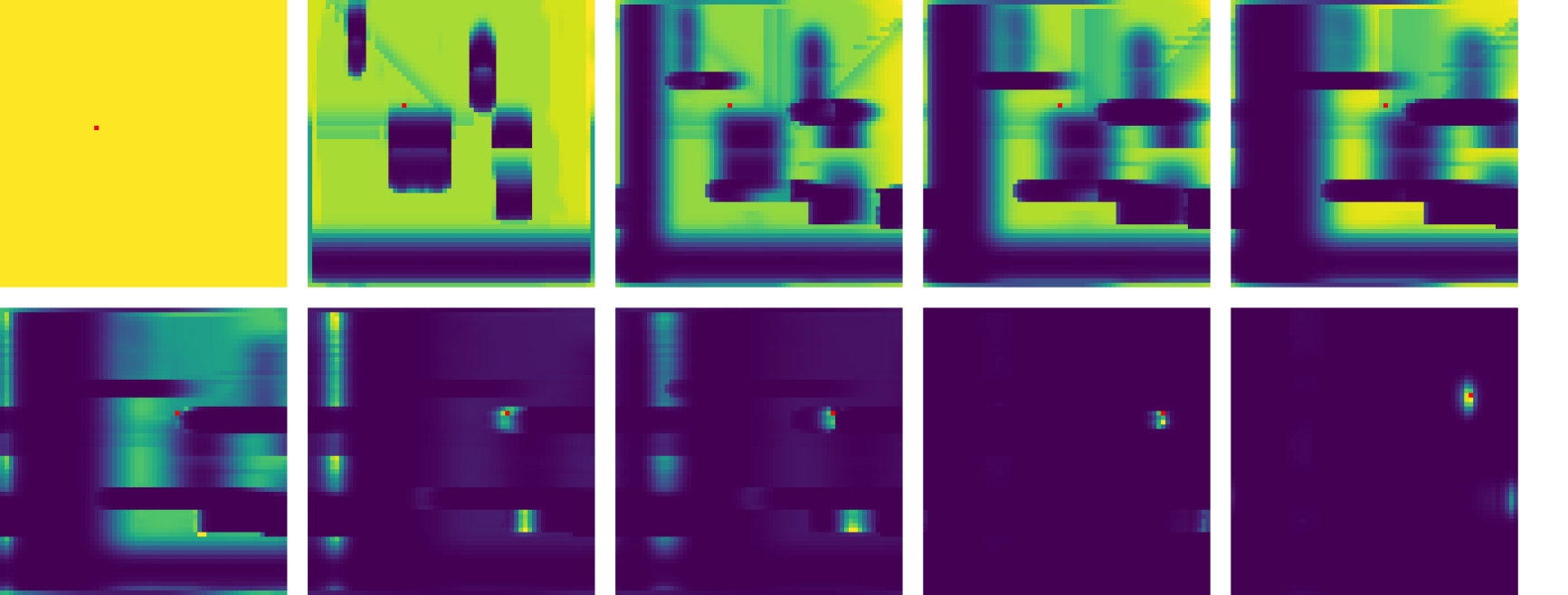}
    \caption{\textbf{Qualitative Results:} We show how the belief change over time for 10 steps. The top-left is the uniform initial belief and the bottom-right is the final belief. The red pixel shows the true location. As the gripper interacts with the objects, the belief becomes less uncertain about the true position.}
    \label{fig:qual_results}
\end{figure*}
We conduct simulation experiments to investigate the localization success of our proposed method and more importantly the generalization capabilities over different size, shape and configurations of the objects. The proposed method is evaluated on two sets of objects: (1) Primitive objects including squares, spheres, capsules, and cylinders. (2) 3DNet objects~\cite{wohlkinger20123dnet} which includes arbitrary shaped objects that are commonly used in daily life. Note that we train the models using the primitive object set and use the 3DNet objects to show that our method can generalize to objects that were not present in the training set. The simulation environment and the object sets can be seen in Figure \ref{fig:sim}. We compare our results against two baselines: (1) a uniform observation model, i.e. $p(o|s) = 1/HW, \forall s\in S$, and (2) a naive version of our observation model which does not use the image encoder module (Figure \ref{fig:network}.a) and directly outputs the likelihood probabilities from the observation encoder module (Figure \ref{fig:network}.b). At anytime during the filtering, the state can be inferred by the \textit{argmax} operation on the belief $\Tilde{s_t} = \textit{argmax}(bel(s_t))$. The $l_1$-norm is used to calculate the error between the predicted state $\Tilde{s_t}$ and the true state $s_t$, $e_t = |s_t - \Tilde{s_t}|$. 

\begin{table}[h]
\centering
\caption{Quantitative Results: Localization Success Rate}
\begin{tabular}{|c | c | c | c|} 
    \hline
    & Uniform & Naive & Bayes Filter (ours)\\ 
    \hline
    Primitive Objects & 18 & 33 & \textbf{93}\\ 
    \hline
    3DNet Objects & 15 & 32 & \textbf{78}\\ 
    \hline 
\end{tabular}
\label{table:results}
\end{table}
The localization is considered to be successful if the error at the end of the episode is less than 4. We generate 100 random configurations for each of the object set and run 10 episodes for each configuration. The starting position of the gripper is also randomly selected. Note that we make sure that the gripper interacts with at least one object on the table. We achieve 93\% success rate for the primitive objects set and 78\% success rate for the 3DNet object set. It is evidenced by Table~\ref{table:results} that our approach outperforms the baselines by a large margin. We observe two failure modes. If the gripper touches an object early in the episode and it localizes well but then its belief becomes uncertain afterwards since it is not receiving any unique observation. The second failure mode is when more than one object has similar size and shape. In this case, there are multiple states that can generate similar observations, therefore, the belief has high probabilities in multiple states. We also show qualitative results in Figure~\ref{fig:qual_results} in which the evolution of the belief over 10 time steps is demonstrated. As can be seen from the figure, the belief gets less uncertain about the correct position of the gripper as it interacts with the objects.

\subsection{Real World Experiments}
To show that our Bayes filter models can be transferred to real world environments, we deployed the models learned in simulation on a real robot. The gripper is attached to a Universal Robot arm (see Figure~\ref{fig:real}) which is programmed to traverse a linear path over the table and objects with constant speed. A Structure depth sensor is placed over the table. The outlier pixels in the image are replaced with the average of neighbor pixels. We applied a low-pass filter on the finger positions to get rid of the noise caused by the bumps on the table and objects. We generated 4 random configuration where we used different objects for each of the configuration. For each configuration, we run 5 episodes from random starting position. To calculate the ground truth states, the position of the gripper is obtained from the forward kinematics and projected in to the camera pixel coordinates. Similar to the simulation experiments, the localization is considered to be successful if the error is under 4. The accuracy of the localization is found to be 90\%. 

\section{CONCLUSIONS}
In this work, we have addressed the problem of tactile localization where the goal is to localize a robotic gripper with respect to an global map of the environment using tactile feedback. We formulate the localization problem as a recursive Bayes filtering problem and learn the filter models from data. The models are implemented as layers of neural networks and trained with gradient descent. A self-supervised and simulation-based data collection procedure is introduced to collect contact interactions between the environment and the gripper. We showed that in addition to successful localization with unseen object configurations, our approach can also localize with novel objects. We also showed that our models can be transferred to real hardware without any domain randomization or retraining. 

The main drawback of our method is the assumption that the gripper does not move objects. To mitigate this problem, we would like to extend the localization formulation to track the objects as well. Another limitation is that the gripper used in this work can operate with low stiffness, thereby,  we can use the joint angles as the tactile observations. However, this might not be true for every gripper. To overcome this, we want to leverage other tactile sensors that can be commonly used. Our future work also includes combining the proposed localization method with policy learning to perform manipulation tasks by jointly learning the localization and planning.

\section*{APPENDIX}
The neural networks are implemented using PyTorch framework \cite{paszke2019pytorch}. All the convolution layers in the image encoder and the likelihood decoder has a kernel size of 3 except the transposed convolutions which has a kernel size 2. The image encoder has 6 layers with the following numbers of filters [16, 16, 32, 32, 64, 64]. The likelihood decoder has a 4 convolution layers and 2 transposed convolution layers with the following numbers of filters [32, 32, 16, 16] and [69, 32], respectively. The observation encoder has 3 layers and each of them has a kernel size of 3. The number of filters are [16, 32, 64]. We collect 600000 trajectories for training set and 60000 trajectories for the validation set. Each trajectory has 140 time steps which results in 84000000 data points. We use Adam optimizer with 0.0003 and 0.001 learning rate for the observation and motion networks. The training for the observation model has batch size of 512 and 300 epochs and the training for the motion model has batch size of 64 and 100 epochs. The observation model is trained in 24 hour and the motion model is trained in 1 hour. 

\addtolength{\textheight}{-5cm}  
\bibliographystyle{IEEEtran} 
\bibliography{references}

\end{document}